\documentclass[a4paper]{article}

\usepackage{INTERSPEECH2022}
\usepackage{amsfonts}
\usepackage{hyperref}
\usepackage{bm}
\usepackage{float}
\usepackage{nicefrac}
\usepackage{subcaption}
\usepackage{multirow}
\usepackage{multicol}
\usepackage{makecell}
\usepackage{adjustbox}
\usepackage{booktabs}
\usepackage{tablefootnote}
\usepackage{xspace}
\usepackage{xcolor} 
\usepackage{algorithm}

\title{Supervision-Guided Codebooks for Masked Prediction in Speech Pre-training}
\name{Chengyi Wang$^{1*}$\thanks{$*$ Equal contribution. The work was done during Chengyi Wang's internship at Microsoft. The authors thank Ashish Arora for help with data preparation in Kaldi.}, Yiming Wang$^{2*}$, Yu Wu$^2$, Sanyuan Chen$^2$, Jinyu Li$^2$, Shujie Liu$^2$, Furu Wei$^2$}

\address{
  $^1$Nankai University \qquad $^2$Microsoft}
\email{cywang@mail.nankai.edu.cn, \{yimingwang,yuwu1,t-schen,jinyli,shujliu,fuwei\}@microsoft.com}

\begin{document}

\maketitle
\begin{abstract}
Recently, masked prediction pre-training has seen remarkable progress in self-supervised learning (SSL) for speech recognition. It usually requires a codebook obtained in an unsupervised way, making it less accurate and difficult to interpret. We propose two supervision-guided codebook generation approaches to improve automatic speech recognition (ASR) performance and also the pre-training efficiency, either through decoding with a hybrid ASR system to generate phoneme-level alignments (named \emph{PBERT}), or performing clustering on the supervised speech features extracted from an end-to-end CTC model (named \emph{CTC clustering}). Both the hybrid and CTC models are trained on the same small amount of labeled speech as used in fine-tuning. Experiments demonstrate significant superiority of our methods to various SSL and self-training baselines, with up to 17.0\% relative WER reduction. Our pre-trained models also show good transferability in a non-ASR speech task. 
  
\end{abstract}
\noindent\textbf{Index Terms}: speech recognition, self-supervised learning, pre-training, self-training, pseudo-labeling, codebook

\section{Introduction}
The automatic speech recognition (ASR) community has witnessed rapid progress over the last decade thanks to the advances in deep learning. However, a drawback of most ASR models is that they require large quantities of labeled training data, which is much harder to come by than unlabeled data, especially for low-resource domains/languages. This drives research in self-supervised learning (SSL), which introduces a pre-training task where a model can be trained with unlabeled data to learn good data representations. The model is then fine-tuned on a relatively smaller amount of labeled data in a conventional supervised way. 

 In terms of how to formulate the loss, existing SSL approaches for speech include directly regressing continuous hidden representations \cite{chung2020generative,baevski2022data2vec}, contrastive prediction \cite{oord2018representation,schneider2019wav2vec,baevski2020wav2vec}, and masked prediction \cite{baevski2020vq,hsu2021hubert}. In the relevant field of natural language computing (NLP), masked language modeling, e.g. BERT \cite{devlin2019bert}, is the dominant pre-training paradigm where masked input tokens are predicted conditioned on the rest of the input sequence. However, unlike text tokens, speech signals are continuous, making it not straightforward to be predictive targets. Also, there is no explicit boundary in speech signals that can be used to chunk speech into linguistically meaningful segments. To tackle the above issues, a codebook, also referred to as quantizer, is required to map continuous features into discrete tokens. Hidden unit BERT (HuBERT) \cite{hsu2021hubert} proposes to create a codebook by offline unsupervised clustering algorithms (e.g., K-means) operating on either pre-computed speech features, or hidden representations from a previously trained HuBERT model. These discrete labels are then used as targets for masked prediction via HuBERT loss, forcing the model to learn meaningful continuous latent representations from unmasked speech input. To obtain good ASR performance, HuBERT usually requires performing two or more stages of iterative pre-training, where in each stage a codebook is created by the last stage HuBERT model. The cluster assignments in the codebook provide consistent information about the underlying acoustic and language content across different speech examples. Although HuBERT demonstrates competitive ASR performance after fine-tuning, there are still problems in the way the codebook is generated: the performance of simple clustering on the unsupervised speech features is not satisfactory and there is no explicit interpretation of what acoustic or language characteristics each cluster corresponds to, which otherwise can potentially be useful or leveraged in other speech tasks/approaches. In addition, usually multiple rounds of pre-training are required to refine the clustering quality, resulting in high time cost. 

To address the issues, we propose to create the codebook from an existing supervised ASR system to improve model pre-training. Specifically, we first train a hybrid ASR system as a teacher model on the available limited labeled data. The hybrid system is able to generate phoneme-level alignments via decoding for all the unlabeled speech. Then we use the phoneme alignments as training targets in HuBERT pre-training. We name this approach as \emph{PBERT} (short for ``phoneme BERT''). Instead of using a hybrid system, we can alternatively train an end-to-end (E2E) CTC model and perform K-means clustering on the supervised speech features (denoted as \emph{CTC clustering}). Compared to clustering on unsupervised speech features as in HuBERT, well-trained ASR systems, either adopting traditional hybrid approaches \cite{hinton2012deep} or recent E2E ones \cite{li2022recent}, can be more preferable in terms of generating more relevant and reliable pre-training targets. In this way, a single round of pre-training suffices to provide a good pre-trained model for fine-tuning. 

We evaluate our method on the LibriSpeech benchmark \cite{librispeech}, where the 100h clean subset is used as labeled data and the other 860h speech as unlabeled data. PBERT and CTC clustering are both superior to the competitive HuBERT baseline. In particular, without a language model (LM), the PBERT outperforms HuBERT by 17.0\% (on \texttt{test-clean}) and 10.0\% (on \texttt{test-other}) relatively with only a single round of pre-training. Pre-training the model for another round or combining PBERT with the self-training technique can further improve the performance. We also evaluate the PBERT on speaker verification task using VoxCeleb dataset. Experiments show that PBERT almost keeps the excellent generalization capability of HuBERT, and significantly outperforms the model initialized with a supervised ASR encoder on the speaker verification task.  


\vspace{-2mm}
\section{Related Work}
\vspace{-3.2mm}
\subsection{HuBERT}
HuBERT is an SSL method benefiting from an offline clustering step to provide target labels for a BERT-like prediction loss \cite{devlin2019bert}. Its backbone is a Transformer encoder \cite{vaswani2017attention}. During pre-training, the Transformer consumes masked acoustic features and the model is optimized to predict the discrete target labels on the masked regions. This forces the model to learn a combined acoustic and language model over the continuous inputs.

HuBERT adopts an iterative re-clustering and re-training process: in the first iteration, the targets are determined by clustering on MFCC features; in the second iteration, a new generation of training targets are created by clustering the latent representations from the model in the first iteration.

\subsection{Self-Training}
Our method shares the similar idea with \emph{self-training} in the sense that they both improve ASR performance with bootstrapping \cite{thomas2013deep,li2019semi,kahn2020self, self_training, xu2020iterative, noise_student}. In self-training, a teacher model is trained on the available labeled data and then the unlabeled set is labeled with this initial model (a.k.a. \emph{pseudo-labeling}).  Finally, a new student model is trained on the combined labeled and pseudo-labeled data. This pseudo-labeling step can be repeated for multiple rounds to refine the labels. Instead of training a supervised student model to generate the whole transcription, our method predicts frame-level phoneme labels only on the masked region. It provides two advantages compared to self-training. First, our method shows excellent transferability in non-ASR tasks (presumably due to its self-supervised nature), enabling solving various speech tasks with one pre-trained model. In addition, the proposed method, like other SSL methods, is complementary to self-training methods for speech recognition (see the section below): substantial gain is observed in the speech recognition task.

\subsection{Combination of SSL and Self-Training}
Previous work \cite{zhang2020pushing,xu2021self_training} shows a synergy between SSL and self-training by employing the pre-trained models in a self-training loop. Specifically, given the labeled dataset $S$ and the unlabeled dataset $U$, it first pre-trains a model on dataset $U$, fine-tunes it on dataset $S$. Then this fine-tuned model is used as the initial teacher model for pseudo-labeling. After that, we can either train a new supervised model or rerun the fine-tuning with the pre-trained model on the mixture of pseudo-labeled and truly labeled data. The combination of SSL and self-training has been shown very effective for extremely low-resource settings where only 10 minutes, 1 hour or 10 hours labeled speech are available. However, on the 100 hours setting, it is less effective and achieves comparable results as an SSL-only baseline \cite{xu2021self_training}. This approach only considers SSL as a better starting point for the self-training. Our method is complementary in that it uses a supervised teacher model to guide pre-training, and the pre-trained model can further be included in the self-training loop.





\begin{small}
\begin{algorithm}[ht]
\caption{Pipeline of our methods}\label{alg:method}
Input: Labeled dataset $S$, Unlabeled dataset $U$.
\begin{enumerate}
\item Train a supervised model $M_0$ on dataset $S$, set $M=M_0$.
\item Generate pseudo-labeled dataset $M(U)$ with $M$.
\item Generate frame-level alignments or K-means clusters $A(U)$ and $A(S)$ with $M$.
\item Pre-train a masked-prediction model $M'$ on dataset  $A(U) \cup A(S)$. 
\item (Optional) set $M=M'$, go to 2.
\item Fine-tune the pre-trained model $M'$ on $S$ or $M(U) \cup S$. 
\item (Optional) Set $M=M'$, generate pseudo-labeled dataset  $M(U)$, go to 5.
\end{enumerate}
\end{algorithm}
\end{small}

\section{Method}
Algorithm \ref{alg:method} shows the training pipeline of our methods. 
Our training starts from a supervised model, either a hybrid (PBERT) or an end-to-end one (CTC clustering), trained on the labeled data. We use this model to perform decoding or K-means clustering on the large scale unlabeled data to obtain frame-level unit/cluster assignments, which are further used as training targets for a masked-prediction model. In our method, the initial supervised model acts like a tokenizer or quantizer to generate discrete training targets. Compared with the assignments from pre-computed speech features or from a self-supervised pre-trained model (e.g., the first-stage HuBERT), an ASR model trained in a supervised way is believed to retain more ASR-related information and thus generate more ASR-specific targets. After fine-tuning on the small labeled dataset, we find only a single round of pre-training can outperform the two-round HuBERT method. Moreover, substantial gain is observed by combining with self-training techniques that use the fine-tuned model as a teacher model.

The performance of PBERT can be further improved by running another round of pre-training. Instead of running clustering on the latent representations, we use the PBERT model as a phoneme predictor where it predicts the corresponding phonemes for each frame (without masking) as the training targets for the next iteration.

\subsection{Supervision-Guided Codebook with a Hybrid Model}
\label{hybrid}
We train a hybrid ASR model on the small amount of labeled data to provide accurate phoneme-level alignments for the unlabeled data. Traditional HMM-GMM models are capable of such a task by providing reliable estimate of the HMM state occupancy for each frame \cite{hinton2012deep}. Using neural networks in place of GMM can usually further boost the performance due to their stronger modeling power. The neural networks can be trained either with a frame-level (e.g., cross-entropy) or a sequence-level (e.g., MMI \cite{bahl1986maximum} and MPE \cite{povey2002minimum}) criterion. We adopt a special MMI criterion named \emph{Lattice-free} MMI (LF-MMI) \cite{povey2016purely} where the denominator graph contains competing hypotheses derived from a phone LM, and the numerator graph encodes all possible sequences of HMM states pertaining to the transcript. In order to yield better alignments, especially for long utterances, the numerator graph is generated in a way so that it is acyclic, directly encoding alternative alignments information from a previous HMM-GMM training stage. The neural acoustic model trained in this way is advantageous over GMM models or neural models trained with other sequence-level criteria, in terms of ASR performance and alignment accuracy \cite{zhang2021lattice}.

Once the hybrid model training is done, a regular decoding process is conducted on the unlabeled data with the trained model, and the state-level alignments are transduced to the phoneme-level alignments based on the predefined HMM topology and context-dependent phone tying rules\footnote{We use a phoneme-based system for illustration purposes. Actually grapheme units, like Chenones \cite{le2019from}, are also applicable to our method.}. The phoneme-level alignments are to be used as targets in PBERT pre-training.

\subsection{Supervision-Guided Codebook with an E2E Model}
Aside from the hybrid model approach, we also propose to train an end-to-end model and use that for K-means clustering. Specifically, given the small labeled dataset, we train a Transformer CTC model. Then we extract the hidden representations for all the speech in the unlabeled set and apply K-means clustering. The cluster assignments are used as the training targets of a HuBERT model. Compared with latent representations in an unsupervised pre-trained model (i.e., the first-stage HuBERT), the CTC latent representations have a higher correlation with meaningful phonemes. So K-means clustering on such representations may yield clusters more helpful for ASR even when the labeled speech data is much less than unlabeled speech. In addition, the training cost of a CTC model is much lower than that of an SSL model (see Section \ref{sec:training_cost}).  

\subsection{Model Architecture}
\label{arch}
Our self-supervised model architecture design is the same as HuBERT. It contains a convolutional feature encoder with a downsampling rate 320 and a 12-layer Transformer encoder. Besides the original convolutional position embedding, we also add a bucket relative position bias following the implementation in \cite{relpos} with 320 embeddings shared across all layers. Each embedding is a scalar that is added to the attention logits and corresponds to a range of possible key-query offsets. The range increases logarithmically up to an offset of 800, beyond which we assign all relative offsets to the same embedding. In our experiments we will show the contribution of the relative bias. 




\section{Experiments}

\subsection{Dataset}
In this paper, we evaluate our method on the LibriSpeech benchmark \cite{librispeech}. It contains 960 hours of training data \texttt{train-960} from read audiobooks with three subsets: \texttt{train-clean-100}, \texttt{train-clean-360} and \texttt{train-other-500}. We consider \texttt{train-clean-100} as the supervised data (leveraging its transcripts), and the rest as the unlabeled data (discarding the transcripts).

\subsection{Setup}

\noindent \textbf{\textit{Pre-training}}\quad For PBERT, we adapt the LibriSpeech recipe from Kaldi \cite{povey2011kaldi} and ensure that only the \texttt{train-clean-100} portion is used in the training pipeline, including HMM-GMM training, i-vector extractor training \cite{saon2013speaker,dehak2011front}, and neural acoustic model training. The acoustic model is a TDNN-F network \cite{povey2018semi} trained with LF-MMI criterion \cite{povey2016purely}. Decoding is performed on \texttt{train-960} with the official 3-gram LM (\texttt{tgsmall}), and then the lattices are rescored with the official 4-gram LM (\texttt{fglarge}). The 1-best path\footnote{The WER on \texttt{train-960} from these 1-best paths is 6.0\%.} is extracted from each lattice and then converted to a phoneme sequence. There are 347 distinct positional-dependent phonemes (including the silence phone) to be used as the training targets.

In CTC clustering, we train the 12-layer Transformer CTC model from scratch on \texttt{train-clean-100}, with 8 GPUs and a batch size of 200 seconds of audio per GPU. After training, we run K-means clustering with 500 clusters on the latent features extracted from the CTC model at the 7-th transformer layer, which has the highest mutual information with phoneme tokens according to the analysis in \cite{hsu2021hubert}.

Our pre-training setting follows LibriSpeech \textsc{Base}\xspace in \cite{hsu2021hubert}. During training, each audio example is cropped to at most 15.6 seconds. The model is pre-trained on 32 GPUs with 87.5 seconds of audio per batch for 400k steps. We use AdamW optimizer \cite{loshchilov2019decoupled} with weight decay 0.01 and $\beta=(0.9,0.98)$. The learning rate ramps up linearly for the first 32k steps and then decays linearly back to 0. The peak learning rates is 5e-4.

\noindent \textbf{\textit{Fine-tuning}}\quad The model is fine-tuned on 8 GPUs with a batch size of 200 seconds of audio per GPU. The convolutional encoder is always fixed. We use Adam optimizer and a \emph{tri-stage} schedule where the learning rate is warmed up for the first 10\% of updates, held constant for the next 40\% and then linearly decayed for the remainder. For evaluation, we use the wav2letter++ \cite{pratap2019wav2Letter} beam search decoder with a beam size 1500 and a 4-gram LM trained on the same text as the other LMs.

\begin{table}[ht]
\caption{Results and comparisons in base model setting. All models only utilize 100h labeled data, and 860 unlabeled data. \label{main_result}}
\vspace{-6mm}
\begin{center}
\begin{adjustbox}{max width=\linewidth}
\begin{tabular}{lccc}
\toprule
\textbf{Model} & \textbf{LM} &  test-clean & test-other \\
\toprule
{\textbf{\textit{Supervised Baselines}}} & & & \\ \hline
\multirow{2}{*}{Transformer CTC} & None  & 8.8 & 26.5 \\   
                                & 4-gram & 5.0 & 16.8 \\ \hline
LF-MMI (Hybrid) & 4-gram & 4.6 & 15.0 \\
\toprule
{\textbf{\textit{Self-supervised Baselines}}} & & & \\ \hline
\multirow{2}{*}{wav2vec 2.0 \cite{baevski2020wav2vec}} & None & 6.1 & 13.3  \\
                                                       & 4-gram & 3.4 & 8.0 \\ \hline
\multirow{2}{*}{HuBERT iter 1 \cite{hsu2021hubert}} & None &     7.4 &  16.2 \\
                                     & 4-gram &  3.9  &  9.5   \\ \hline
\multirow{2}{*}{HuBERT iter 2 \cite{hsu2021hubert}} & None & 5.9 & 13.0   \\
                                             & 4-gram  & 3.4 & 8.1\\ 
                                          
\cmidrule(lr){2-4}
\multirow{2}{*}{\quad + rel bias} & None & 5.7 & 12.3 \\
                                   & 4-gram & 3.4 & 8.1 \\ \hline
 \multirow{2}{*}{Random-codebook \cite{chiu2022self} + rel bias} & None & 6.9  & 14.6 \\
                                           & 4-gram & 3.7  &  9.0 \\ \hline
                                           
   \multirow{2}{*}{WavLM + rel bias \cite{chen2021wavlm}}  & None & 5.7 & 12.0\\       
  & 4-gram & 3.4 & 7.7 \\                                 
\toprule
{\textbf{\textit{Self-training Baselines}}} & & & \\ \hline
Self Training (ST) \cite{self_training}  &  GCNN &  5.8 & 20.1 \\ \hline
IPL \cite{xu2020iterative}  &  4-gram + Trans.   & 5.6 & 9.0 \\ \hline
Noisy Student \cite{noise_student}  & LSTM & 4.2 & 8.6 \\ \hline
\multirow{2}{*}{self-training (Ours)} & None & 4.9 & 14.4 \\
                                     & 4-gram & 3.5 & 9.7 \\
\cmidrule(lr){2-4}
\multirow{2}{*}{\quad + 2nd iteration} & None & 4.3 & 11.0 \\ 
                                     & 4-gram & 3.3 & 8.4 \\

\toprule
{\textbf{\textit{Our Methods}}} & & & \\ \hline

\multirow{2}{*}{PBERT} & None & 4.9 & 11.7 \\
                       & 4-gram & \textbf{3.1}  & 7.7 \\
\cmidrule(lr){2-4}
\multirow{2}{*}{\quad + rel bias} & None & 4.7 & 11.2 \\
                                 & 4-gram & \textbf{3.1} & 7.5 \\
\cmidrule(lr){2-4}
\multirow{2}{*}{\quad\quad + 2nd iteration} & None & 4.7 &  10.7 \\
                                   & 4-gram & \textbf{3.1}  &  7.3 \\
\cmidrule(lr){2-4}
\multirow{2}{*}{\quad\quad + self-training} & None & \textbf{4.2} &  \textbf{9.5} \\
                                   & 4-gram & \textbf{3.1} &  \textbf{7.2} \\ \hline
\multirow{2}{*}{CTC clustering + rel bias} & None & 5.2 & 11.4 \\ 
       & 4-gram &  3.2  &  7.4 \\ \hline
\multirow{2}{*}{Ground-truth phones + rel bias} & None & \it\textcolor{gray}{4.5} & \it\textcolor{gray}{10.0} \\
                                     & 4-gram & \it\textcolor{gray}{3.1} & \it\textcolor{gray}{6.8} \\
\bottomrule
\end{tabular}
\end{adjustbox}
\vspace{-8mm}
\end{center}
\end{table}

\subsection{LibriSpeech 100-860 Results}
Table \ref{main_result} presents the results compared against the self-supervised and self-training baselines. The hybrid model and the CTC model that we used to generate codebooks are listed as the supervised baselines.

\noindent\textbf{\textit{Self-supervised learning}} \quad We compare with four self-supervised baselines. Wav2vec2.0 is a contrastive learning SSL method. Others are masked prediction pre-training methods: HuBERT predicts the discrete targets unsupervisely derived from original speech features or latent representations; a random codebook is proposed in \cite{chiu2022self} where they use a random projection quantizer to generate discrete labels. WavLM introduces a data augmentation method to improve HuBERT. We re-implement the HuBERT, Random-projection and WavLM baselines with the same model size equipped with relative position bias for a fair comparison.

Our proposed methods rely on supervised models to provide high quality  training targets for pre-training and perform well on both test sets. Without an LM, PBERT obtains a relative 17.0/10.0\%  WER reduction over HuBERT. Relative positional embedding can further bring a 2.6\% relative gain on \texttt{test-other}. The CTC clustering method is slightly worse than PBERT since the pre-training targets are obtained by K-means clustering and thus less interpretable than phonemes. Compared to other baselines, PBERT with pseudo phoneme labels is significantly better on \texttt{test-clean}, indicating the supervision-guided codebook with \texttt{train-clean-100} could provide a better pre-training objective for in-domain (clean) data.  Moreover, we train a PBERT on the force-aligned phoneme labels\footnote{The phoneme labels are obtained by force-aligning labeled \texttt{train-960} with a hybrid model trained on \texttt{train-960}.}, which is denoted as ``Ground-truth phones + rel bias'' in the table and can be considered as the upper bound of our methods. 

In addition, PBERT could perform iterative pre-training like HuBERT (denoted as ``+ 2nd iteration'' in Table \ref{main_result}). We use the pre-trained PBERT model as a phoneme predictor to generate new frame-level pseudo targets for all training data. Then we run another round of pre-training with the new phoneme targets. The new pre-trained model is also fine-tuned on \texttt{train-clean-100}. This method improves the WER on \texttt{test-other} from 11.2\% to 10.7\%, but has no gain on \texttt{test-clean}. This implies that the second iteration of pre-training can generate better phoneme labels for unlabeled noisy data, while for clean data label quality is not improved. One possible explanation is that the hybrid model, trained with in-domain data only, does not perform well for out-domain data label generation, and the first round pre-trained model improves the model robustness on the out-domain data (i.e., \texttt{test-other}).  

\noindent\textbf{\textit{Self-training}} \quad For self-training baselines, we list the literature results \cite{self_training, noise_student, xu2020iterative} under this 100h-860h setting, and also run our own experiments with Transformer models (w/o relative bias).  For the first iteration, we use the hybrid model as the initial teacher and train a Transformer CTC model from scratch as the student model. Then we regard this Transformer model as a new teacher and generate another round of pseudo-labels with 4-gram LM shallow fusion. For both iterations, we do not perform filtering on pseudo-labeled examples and directly mix the 860h pseudo-labeled data and the 100h truly labeled data.
 
Self-training is a strong baseline, which performs better than HuBERT on \texttt{test-clean} but worse on the noisy \texttt{test-other}. This is due to our labeled training data having the same domain as the clean set and thus the pseudo-labels generated by the teacher model for the clean set are of higher quality than the noisy set. Our proposed method is compatible with self-training.  We can see that one round of self-training could further bring WER down to 4.2/9.5\% without LM, and 3.1/7.2\% with LM (better than the self-training-only baseline by 6.1/14.3\% respectively). An interesting phenomenon is that the self-training only provides limited gain in the 4-gram LM fusion setting. Our explanation is the pseudo-labeled data are created by decoding with LM, so the LM knowledge has been distilled into the fine-tuned model. Therefore, the 4-gram LM cannot contribute a lot to the final results. 

\subsection{Non-ASR Task Transfer}
Existing works have demonstrated that SSL enjoys outstanding generalizability and re-usability across various downstream scenarios \cite{superb, chen2021wavlm}. An interesting question is whether our model maintains the advantage of SSL if the phoneme is the target in pre-training. To answer this question, we fine-tune our models on VoxCeleb2 \cite{chung2018voxceleb2}, and evaluate its performance on the three official trial lists Vox1-O, Vox1-E and Vox1-H, following the pipeline proposed in \cite{chen2021large}.  
    \begin{small}
\begin{table}[ht]
    \centering
    \vspace{-1mm}
    \caption{Equal error rates (EERs) on VoxCeleb 1 speaker verification test set. 
    }
        \vspace{-3mm}

    \label{tab:asv}
    \begin{tabular}{l|ccc} 
         \toprule 
         \multirow{2}{*}{Model} & \multicolumn{3}{c}{EER (\%)} \\
         & Vox1-O & Vox1-E & Vox1-H \\
         \hline
         FBank & 1.01 & 1.24 & 2.32 \\
         \hline
        ASR Encoder & 1.159 & 1.256 & 2.434  \\
        wav2vec 2.0  & 0.973 & 0.933 & 1.831 \\
        HuBERT & 0.84 & 0.879 & 1.726 \\
        PBERT  & 0.867 & 0.918 & 1.776 \\
        
         \bottomrule
    \end{tabular}
    \vspace{-1mm}
\end{table}
        \end{small}

Table \ref{tab:asv} shows the results of different pre-trained encoders being combined with the downstream model ECAPA-TDNN. ``ASR encoder'' means we pre-train a CTC model with labeled \texttt{train-960}, and feed the ASR encoder outputs to the downstream model to obtain speaker embeddings. We can observe that the ASR encoder does not enjoy a good generalizability, while all the pre-trained models outperform ECAPA-TDNN + fbank feature significantly. PBERT is slightly worse than HuBERT on the speaker verification with less than $3\%$ degradation. One possible explanation is that the phoneme labels, which are obtained from an ASR model, might lose speaker dependent characteristics compared to HuBERT.  

\subsection{Training Cost of Different Codebooks}
\label{sec:training_cost}

We compare the training costs of different codebook generation approaches. HuBERT and CTC clustering method are implemented with PyTorch and trained on Tesla V100 GPUs. They require 866.7 GPU hours and 53.3 GPU hours respectively. The training cost of CTC clustering is  93.8\% less than HuBERT. The training of our hybrid system is with Kaldi on Tesla K10 GPUs, and only 37.9 GPU hours are required\footnote{The alignment model (HMM-GMM) training is still performed on CPUs. However, recent work \cite{zhang2021lattice} has shown that an HMM-DNN model can serve as a better alignment model that runs on GPUs.}.

\section{Conclusion}
In this paper, we propose to use a supervision-guided codebook for speech pre-training via masked prediction. The codebook is obtained by either training a hybrid ASR model or an E2E ASR model with limited supervised data. The codebook enjoys better interpretablity, lower training cost, and leads to stronger downstream speech recognition performance. Experiment results show that our model significantly outperforms the HuBERT baseline on the LibriSpeech benchmark, and is compatible with self-training methods.

\bibliographystyle{IEEEtran}

\bibliography{mybib}


\end{document}